\documentclass[10pt,twocolumn,letterpaper]{article}

\usepackage{iccv}
\usepackage{times}
\usepackage{epsfig}
\usepackage{graphicx}
\usepackage{amsmath}
\usepackage{amssymb}

\usepackage{tikz}
\usepackage{comment}
\usepackage{amsmath,amssymb} % define this before the line numbering.
\usepackage{color}

\usepackage{multicol}
\usepackage{multirow}
\usepackage{booktabs}

\newlength\savewidth\newcommand\shline{\noalign{\global\savewidth\arrayrulewidth
  \global\arrayrulewidth 1pt}\hline\noalign{\global\arrayrulewidth\savewidth}}
  
\DeclareMathOperator{\avg}{avg}

\usepackage[breaklinks=true,bookmarks=false]{hyperref}

\iccvfinalcopy % *** Uncomment this line for the final submission

\ificcvfinal\fi

\begin{document}

%%%%%%%%% TITLE
\title{Polarimetric Information for Multi-Modal 6D Pose Estimation of Photometrically Challenging Objects with Limited Data}

\author{Patrick Ruhkamp$^{1,2}$
\and
Daoyi Gao$^{1}$
\and
HyunJun Jung$^{1}$
\and
Nassir Navab$^{1}$
\and
Benjamin Busam$^{1,2}$
\and
$^{1}$ TU Munich, $^{2}$ 3Dwe.ai\\
{\tt\small p.ruhkamp \{...\} b.busam@tum.de}
}

\maketitle
\ificcvfinal\thispagestyle{empty}\fi

%%%%%%%%% ABSTRACT
\begin{abstract}
   6D pose estimation pipelines that rely on RGB-only or RGB-D data show limitations for photometrically challenging objects with e.g. textureless surfaces, reflections or transparency. 
   A supervised learning-based method utilising complementary polarisation information as input modality is proposed to overcome such limitations. This supervised approach is then extended to a self-supervised paradigm by leveraging physical characteristics of polarised light, thus eliminating the need for annotated real data. The methods achieve significant advancements in pose estimation by leveraging geometric information from polarised light and incorporating shape priors and invertible physical constraints. 
\end{abstract}

%%%%%%%%% BODY TEXT
\section{Introduction}
We address limitations of 6D object pose estimation using depth sensor data affected by inaccuracies from artifacts like multi-path interference, ambient light, and transparency~\cite{wang2021occlusion}. Utilising polarised light's geometric information for pose estimation has shown promising results, outperforming RGB-only and RGB-D methods~\cite{gao2021polarimetric}. However, obtaining extensive annotated datasets is challenging~\cite{PhoCal,jung2022housecat6d}. 
We investigate the effectiveness of polarimetric data in improving 6D object pose estimation through supervised learning. Additionally, we propose a physics-induced polarimetric self-supervision scheme to alleviate the need for large annotated real datasets~\cite{ruhkamp2023s2p3}.

\section{Related Work}

\begin{figure}[!b]
      \centering
      \includegraphics[width=0.7\columnwidth]{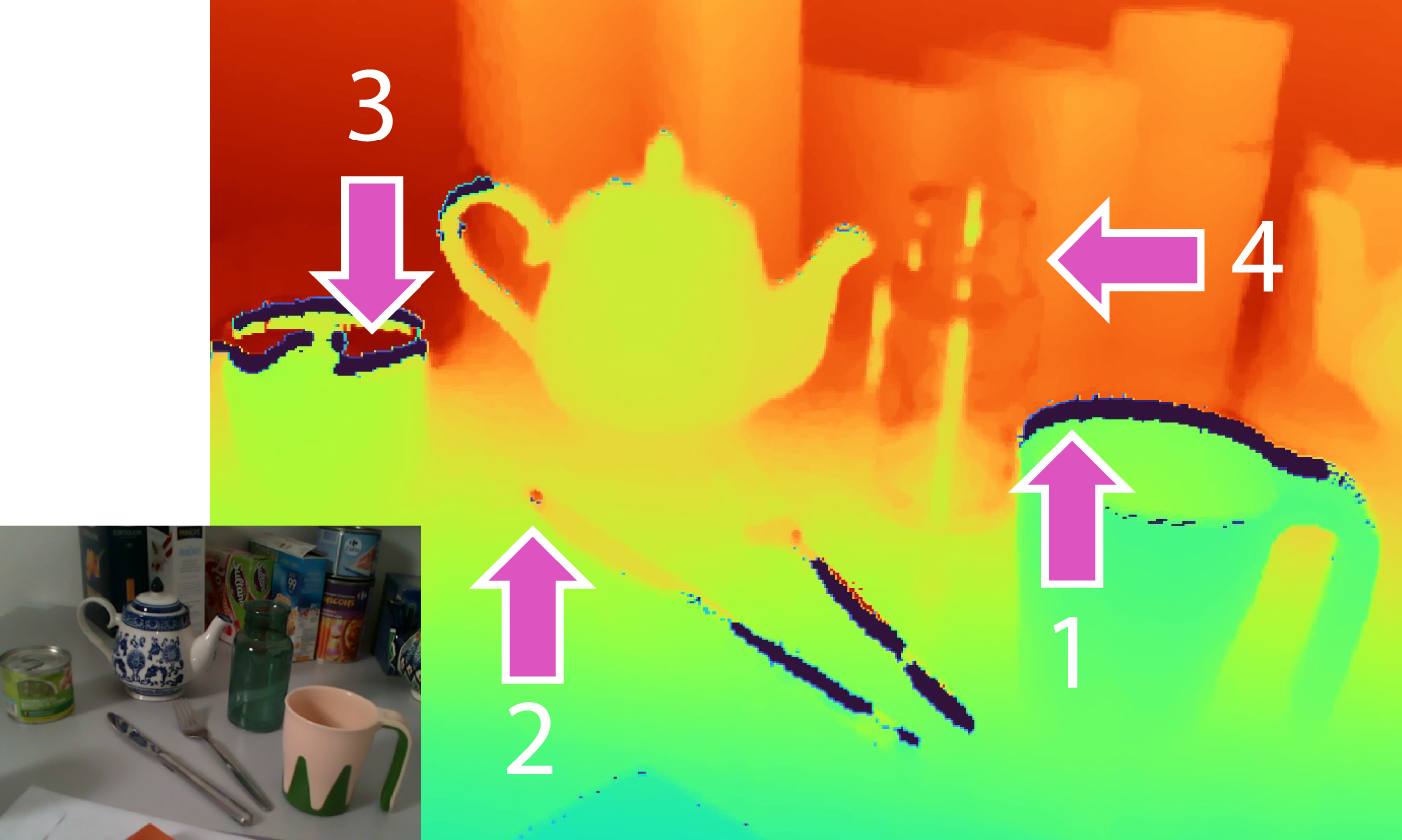}
      \caption{\textbf{Challenges in Depth Estimation for Household Objects. } Reflective boundaries (1,3) and strong reflections (2,3) cause depth miscalculations, while semi-transparent objects (4) become partially invisible to the depth sensor.
      }
      \label{fig:rgbd_issues}
\end{figure}

\noindent\textbf{Polarimetric Imaging. }
Early studies on shape from polarisation (SfP) focus on estimating surface normals and depth information by analysing the relationship between polarisation and object surfaces. However, these works mainly operate in controlled lab environments~\cite{atkinson2006recovery,garcia2015surface,smith2018height,yu2017shape}. Monocular polarisation images are commonly used, but SfP can also utilise multiple views for depth estimation~\cite{atkinson2005multi,cui2017polarimetric}. Polarimetric images are combined with photometric information from stereo~\cite{atkinson2017polarisation} or monocular RGB~\cite{zhu2019depth} to enhance depth predictions. Additionally, polarised light improves noisy depth maps obtained from other sensors~\cite{kadambi2017depth}. Recent work by Ba et al.~\cite{ba2020deep} employs neural networks to predict surface normals by leveraging plausible cues extracted from polarimetric images, demonstrating the disambiguation capability of such cues in SfP. Inspired by these findings, our approach incorporates shape priors from physical properties extracted from polarised light to complement pose estimation.

\noindent\textbf{6D Object Pose Estimation. }
Dense correspondence-based methods have gained popularity for 6D object pose estimation~\cite{zakharov2019dpod,hodan2020epos,li2019cdpn,park2019pix2pose,shugurov2021dpodv2}. These methods involve training neural networks to predict 2D-3D correspondences between object pixels and their corresponding 3D locations on the object's surface. The correspondences are then used with PnP+RANSAC, the Umeyama algorithm, or direct regression to compute the 6D object pose. ZebraPose~\cite{su2022zebrapose} introduces hierarchical feature representations, and zero-shot methods are being explored for 6D pose estimation~\cite{shugurov2022osop}. However, many correspondence-based methods suffer from computationally expensive post-processing steps in RANSAC-based pose solvers. GDR-Net~\cite{wang2021gdr} and its follower SO-Pose~\cite{di2021so} address this limitation by employing learning-based MLP networks to directly predict the target pose from dense correspondences, thereby improving computational efficiency. In our work, we build upon these findings to directly regress the object pose.

\noindent\textbf{Geometric Depth Information. }
Several methods, including FFB6D~\cite{He_2021_CVPR}, Uni6D~\cite{jiang2022uni6d}, ESA6D~\cite{mo2022es6d}, FS6D~\cite{he2022fs6d}, and DGECN~\cite{cao2022dgecn}, incorporate depth information into their prediction pipelines. However, these approaches heavily rely on depth quality, which is problematic for photometrically challenging objects~\cite{gao2021polarimetric}. Geometric cues derived from polarization can potentially mitigate these issues.

\noindent\textbf{Polarimetric 6D Pose Prediction. }
Recently published annotated datasets for real-world polarimetric category-level~\cite{PhoCal} and instance-level~\cite{gao2021polarimetric} 6D pose estimation enable the study of methods using this unexplored imaging modality. PPP-Net~\cite{gao2021polarimetric} investigates the advantages of using polarisation for object pose estimation and proposes a hybrid pipeline that combines physical model cues with learning.
Unlike previous methods such as Self6D~\cite{wang2020self6d} and Self6D++~\cite{wang2021occlusion}, we leverage polarimetric images and an extended differentiable renderer to incorporate geometric representations and enable self-supervision based on our invertible physical model.

\section{Polarimetric Physical Model}
\label{polarization}
RGB-D sensors often rely on active illumination for depth measurement. However, they are susceptible to photometric challenges such as translucency and reflections, leading to erroneous depth estimates. Figure \ref{fig:rgbd_issues} demonstrates such depth artifacts. We instead leverage the surface normals obtained from polarisation for 6D object pose estimation, where accurate 3D information is crucial.
When unpolarised light passes through a linear polariser or is reflected at Brewster's angle from a surface, it becomes polarised. The combination of specular and diffuse reflection results in a partially polarised reflected light, which carries information about the surface properties. We propose to use surface normals obtained from polarisation to overcome the photometric challenges faced by RGB-D sensors. 

\paragraph{Image Formation Model.}
The polarisation image formation model~\cite{fliessbach2012elektrodynamik} considers the degree of polarization (DoP) $\rho$ and angle of polarisation (AoP) $\phi$ of the incoming light. By analysing the oscillation state of light captured by polarisation filters at different angles $\varphi_{pol}$, the polarised intensities can be computed using:
$I_{\varphi_{pol}} = I_{un} \cdot \ (1+\rho \  \cos(2(\phi - \varphi_{pol})))$. 
We find $\varphi$ and $\rho$ from the over-determined system of linear equations using linear least squares. Depending on the surface properties, AoP is calculated as:
\begin{align}
    \left\{
    \begin{array}{lll}
        \phi_{d} [\pi] &= \alpha \ &\text{for diffuse reflection}\\
        \phi_{s} [\pi] &= \alpha - \frac{\pi}{2} \ &\text{for specular reflection,}
    \end{array}
    \right.
    \label{eqn:aolp}
\end{align}
where $[\pi]$ indicates the $\pi$-ambiguity and $\alpha$ is the azimuth angle of the surface normal $\textbf{n}$.
We can further relate the viewing angle $\theta \in [0, \pi/2]$ to the degree of polarisation by considering Fresnel coefficients, thus DoP is similarly given by~\cite{atkinson2006recovery}
\begin{align}
    \left\{
    \begin{array}{l}
        \rho_{d} = \frac
            {(\eta-1/\eta)^{2}\sin^{2}(\theta)}
            {2+2\eta^{2}-(\eta+1/\eta)^{2}\sin^{2}(\theta)+4\cos(\theta)\sqrt{\eta^{2}-\sin^{2}(\theta)}}
            \\
            \\
        \rho_{s} =  \frac
            {2\sin^{2}(\theta)\cos(\theta)\sqrt{\eta^{2}-\sin^{2}(\theta)}}
            {\eta^{2}-\sin^{2}(\theta)- \eta^{2}\sin^{2}(\theta) +2\sin^{4}(\theta),}
    \end{array}
    \right.
    \label{eqn:dolp}
\end{align}
with the refractive index of the observed object material $\eta$.
Solving equation~\ref{eqn:dolp} for $\theta$, we retrieve three solutions $\theta_d,\theta_{s1},\theta_{s2}$, one for the diffuse case and two for the specular case. For each of the cases, we can now find the 3D orientation of the surface by calculating the surface normals $
    \mathbf{n} = \left(
    \cos{\alpha}\sin{\theta}, 
    \sin{\alpha}\sin{\theta},
    \cos{\theta}
    \right)^{\text{T}}.
    $
We use these plausible normals $\mathbf{n}_{d}, \mathbf{n}_{s1}, \mathbf{n}_{s2}$ as physical priors per pixel as input to the neural network.

\paragraph{Invertible Physical Model.}
The inverted physical model takes a rendered object surface normal map and derives the analytical polarimetric parameters, considering different reflection properties. The viewing angle $\theta_v$ is obtained from the dot product between the rendered surface normal map and the viewing vector $\mathbf{v}$. The analytical DoP $\hat{\rho}$ is derived for diffuse and specular reflection cases using Equation \ref{eqn:dolp}. 
This inverted physical model enables us to optimise the model using object shape cues, which is more robust in challenging lighting conditions compared to active depth sensors.

\section{Polarimetric 6D Object Pose Estimation}

\begin{figure*}[!hpt]
      \centering
      \includegraphics[width=0.8\textwidth]{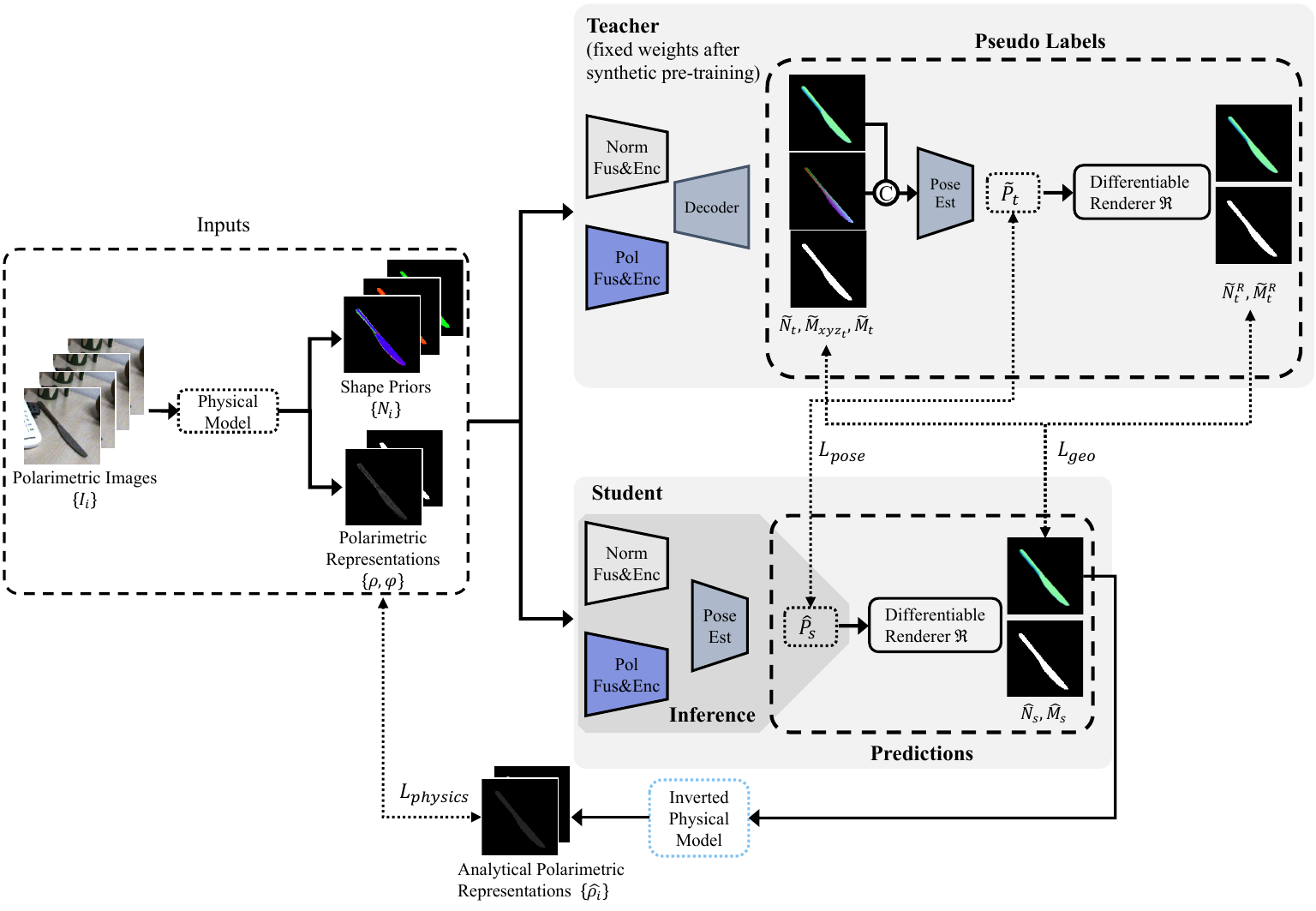}
      \caption{\textbf{Training Scheme Overview. }The upper branch, up until the pose estimation, is equivalent to the supervised training scheme. The training scheme utilises four polarisation images captured at different filter angles along with polarimetric and geometric representations derived from the physical model. The student network is optimised using pseudo labels ($L_{pseudo}$) generated by the teacher and by minimising the discrepancy between $\rho$ from the physical model and $\hat{\rho}$ from the inverted physical model ($L_{physics}$). During inference, the lightweight student network predicts direct pose estimates (dark gray background).
      }
      \label{fig:pipeline}
\end{figure*}

Polarimetric images provide additional geometric information that improves the accuracy of 6D object pose prediction~\cite{gao2021polarimetric} (cf. Inputs in Fig.\ref{fig:pipeline}). We explore its effectiveness in supervised training on real data and propose a self-supervised approach leveraging the analytical physical model, thus eliminating the need for annotated real data.
Here, the supervised network is only pre-trained on synthetic data. The polarimetric real data is leveraged as self-supervision signal against the analytically derived polarimetric representation of the output of a differentiable renderer based on the predicted pose (cf. Fig.\ref{fig:pipeline}).

\begin{table*}[!hpt]
\centering
\caption{\textbf{Supervised Modalities Evaluation. }
Impact of input-output modalities on supervised pose estimation (ADD).
}
\footnotesize
\resizebox{0.6\textwidth}{!}{
\begin{tabular}{l|c|c|c|c|c|c|c|c|c|c|c|c}
\shline
\multirow{2}{*}{Obj.} & \multirow{2}{*}{\shortstack[c]{Photo.\\Chall.}} & \multicolumn{3}{c|}{Input Modalities} & \multicolumn{2}{c|}{Outputs} & \multicolumn{5}{c|}{Normal Metrics} & Pose \\
& \multicolumn{1}{c|}{} & \multicolumn{1}{c}{RGB} & \multicolumn{1}{c}{Pol RGB} & \multicolumn{1}{c|}{Phy. N} & \multicolumn{1}{c}{N} & \multicolumn{1}{c|}{NOCS} & \multicolumn{1}{c}{mean$\downarrow$} & \multicolumn{1}{c}{med.$\downarrow$} & \multicolumn{1}{c|}{11.25$^\circ$ $\uparrow$} & \multicolumn{1}{c|}{22.5$^\circ$ $\uparrow$} & \multicolumn{1}{c|}{30$^\circ$ $\uparrow$} & \multicolumn{1}{c}{ADD}\\
\hline
\multirow{4}{*}{Cup} & \multirow{4}{*}{} & \checkmark & & & & \checkmark & - & - & - & - & - & 91.1 \\
                                                                   & & & \checkmark & & & \checkmark & - & - & - & - & - & 91.3 \\
                                                                   & & & \checkmark & & \checkmark & \checkmark & 7.3 & 5.5 & 86.2 & 96.1 & 97.9 & 91.3 \\
                                                                   & & & \checkmark & \checkmark & \checkmark & \checkmark & \textbf{4.5} & \textbf{3.5} & \textbf{94.7} & \textbf{99.1} & \textbf{99.6} &  \textbf{97.2} \\
\hline
\multirow{4}{*}{Fork} & \multirow{4}{*}{\texttt{$\dagger\dagger$}} & \checkmark & & & & \checkmark & - & - & - & - & - & 85.4 \\
                                                                   & & & \checkmark & & & \checkmark & - & - & - & - & - & 86.1 \\
                                                                   & & & \checkmark & & \checkmark & \checkmark & 11.0 & 7.3 & 72.6 & 90.7 & 93.9 & 92.9\\
                                                                   & & & \checkmark & \checkmark & \checkmark & \checkmark & \textbf{6.5} & \textbf{4.3} & \textbf{87.6} & \textbf{95.9} & \textbf{97.6} &  \textbf{95.9} \\
\shline
\end{tabular}}
\label{tab:ablation}
\end{table*}

\paragraph{Supervised Learning.}
\label{sec:loss}
The overall objective for the supervised learning is composed of both geometrical features learning and pose optimisation~\cite{gao2021polarimetric}, as: $\mathcal{L} = \mathcal{L}_{pose} + \mathcal{L}_{geo}$, with:
    $\mathcal{L}_{pose} = \mathcal{L}_{R} + \mathcal{L}_{center} + \mathcal{L}_{z}$ and  
    $\mathcal{L}_{geo} = \mathcal{L}_{mask} + \mathcal{L}_{normals} + \mathcal{L}_{xyz}$.
Specifically, we employ separate loss terms for given ground truth rotation $\mathbf{R}$, $(\delta_x, \delta_y)$ and $\delta_z$ as:
\begin{equation}
    \begin{cases}
        \mathcal{L}_{R} &= \underset{\mathbf{x} \in \mathcal{M}}{\avg} \| {\mathbf{R}} \mathbf{x} - \hat{\mathbf{R}} \mathbf{x} \|_1 \\
        \mathcal{L}_{center} &= \| ({\delta}_x - \hat{\delta}_x,  {\delta}_y - \hat{\delta}_y) \|_1 \\
        \mathcal{L}_{z} &= \| {\delta}_z - \hat{\delta}_z \|_1
    \end{cases}
    ,
\label{eq:loss_pose_detail}
\end{equation}
where $\hat{\bullet}$ denotes prediction. For symmetrical objects, the rotation loss is calculated based on the smallest loss from all possible ground-truth rotations under symmetry.

\begin{table*}[!t]
\centering
\caption{\textbf{Supervised 6D Pose Benchmark comparisons. } 
We compare our method against recent RGB-D (FFB6D~\cite{He_2021_CVPR}) and RGB-only (GDR-Net~\cite{wang2021gdr}) methods on challenging objects with varying photometric conditions ($\dagger$) and depth map quality (good: $+$ to low:$-$). We report the average recall of ADD(-S) for evaluation.
}
\footnotesize
\resizebox{0.6\textwidth}{!}{
\begin{tabular}{l|c|c|c|c|c|c|c|c|c||c|c}
\shline
\multirow{2}{*}{Object} & \multirow{2}{*}{\shortstack[c]{Photo.\\Chall.}} & \multicolumn{5}{c|}{\multirow{1}{*}{Properties}} & \multirow{2}{*}{\shortstack[c]{Depth\\Quality}} & \multicolumn{2}{c||}{RGB-D Split} & \multicolumn{2}{c}{RGB Split} \\
&& \multicolumn{1}{|c}{\multirow{1}{*}{Refl.}} & \multicolumn{1}{c}{\multirow{1}{*}{Metal}} & \multicolumn{1}{c}{\multirow{1}{*}{Textureless}} & \multicolumn{1}{c}{\multirow{1}{*}{Transp.}} & \multicolumn{1}{c|}{\multirow{1}{*}{Symm.}}
&& \multicolumn{1}{c}{\multirow{1}{*}{FFB6D}} & \multicolumn{1}{c||}{\multirow{1}{*}{\textbf{Ours}}} & \multicolumn{1}{c}{\multirow{1}{*}{GDR}} & \multicolumn{1}{c}{\multirow{1}{*}{\textbf{Ours}}}\\
\cline{1-10}\cline{11-12}
Cup & &&&&&& \texttt{(+)}& \textbf{99.4} & 98.1 & 96.7 & \textbf{97.2}\\
Teapot & \texttt{$\dagger$} & (*) && &  &  &\texttt{++} & 86.8 & \textbf{94.2} & 99.0 & \textbf{99.9}\\
\hline
Can & \texttt{$\dagger$} & * & * & & & & \texttt{-} & 80.4 & \textbf{99.7} & 96.5 & \textbf{98.4}\\
Fork & \texttt{$\dagger\dagger$} & * & * & * &  &  & \texttt{--} & 37.0 & \textbf{72.4} & 86.6 & \textbf{95.9}\\
Knife & \texttt{$\dagger\dagger$} & * & * & * & & &  \texttt{---} & 36.7 & \textbf{87.2} & 92.6 & \textbf{96.4}\\
\hline
Bottle & \texttt{$\dagger\dagger\dagger$} & * & & * & *  & * &  \texttt{None} & 61.5 & \textbf{93.6} & 94.4 & \textbf{97.5}\\
\hline
Mean & &&&&&&& 67.0 & \textbf{90.9} & 94.3 & \textbf{97.6} \\
\shline
\end{tabular}
}
\label{tab:baseline:rgbd}
\end{table*}

To learn the intermediate geometrical features, we employ additional losses with masks ${\mathbf{M}}$: 
\begin{equation}
    \begin{cases}
        \mathcal{L}_{mask} &= \| {\mathbf{M}}  - \hat{\mathbf{M}} \|_1 \\
        \mathcal{L}_{xyz} &= \mathbf{M} \odot \| \mathbf{M}_{xyz} - \hat{\mathbf{M}}_{xyz} \|_1 \\
        \mathcal{L}_{normal} &= 1- \langle\mathbf{n}, \hat{\mathbf{n}}\rangle
    \end{cases}
\label{eq:loss_geo_detail}
\end{equation}
\noindent where $\odot$ indicates the Hadamard product of element-wise multiplication, and $\langle\bullet,\bullet\rangle$ denotes the dot product.

\paragraph{Self-Supervised Learning.}
The self-supervised network (cf. Fig.~\ref{fig:pipeline}) consists of a teacher network and a lightweight student network. Pre-trained on synthetic data, the networks use pseudo-labels from the teacher to guide self-supervised learning of the student on real data. Our approach enhances and modifies established student-teacher training schemes for 6D object pose estimation~\cite{wang2021occlusion}, with detailed explanations and justifications in the following.

The teacher network (based on PPP-Net~\cite{gao2021polarimetric} as in the supervised scenario above) takes polarimetric intensities and shape priors as inputs. It predicts object mask $\mathbf{\Tilde{M}_t}$, object normal map $\mathbf{\Tilde{N}t}$, and dense correspondences map $\mathbf{\Tilde{M}{{xyz}_t}}$. It estimates rotation and translation vectors to represent the pose as $\mathbf{\Tilde{P}_t} = [\mathbf{\Tilde{R}_t} \mid \mathbf{\Tilde{t}_t}]$. The teacher network uses a differentiable renderer to generate pseudo labels, including object mask $\mathbf{\Tilde{M}^R_t}$ and object normal map $\mathbf{\Tilde{N}^R_t}$.

The lightweight student network directly predicts the pose $\mathbf{\hat{P}_s}$ without a geometric decoder. It is optimised using pseudo labels from the teacher network and generates an object normal map $\mathbf{\hat{N}_s}$ and mask $\mathbf{\hat{M}_s}$ through rendering based on the predicted pose.

Our training scheme combines pseudo label learning and physics-induced self-supervision using polarimetric images. The pseudo label loss ($\mathcal{L}_{\text{pseudo}}$) transfers knowledge from the teacher network to the student network, while the physical loss term ($\mathcal{L}_{\text{physics}}$) optimizes the student's prediction using raw polarisation data and the inverted physical model. These terms contribute to the overall loss ($\mathcal{L}$).

The pseudo label loss ($\mathcal{L}_{pseudo}$) is formulated as:
$\mathcal{L}_{pseudo} = \lambda_1\mathcal{L}_{pose} + \mathcal{L}_{geo}$,
where $\mathcal{L}_{pose}$ measures the discrepancy between the predicted pose of the student network and the pseudo ground truth pose provided by the teacher network. The $\mathcal{L}_{geo}$ term regularises the predicted mask and normal map by comparing them with the rendered mask and normal map. The weighting factor $\lambda_1$ depends on the alignment between the predicted shape and pose.

The physics-induced self-supervision loss ($\mathcal{L}_{physics}$) is formulated as:
$\mathcal{L}_{physics} = \underset{\mathbf{x} \in { \hat{\rho_{d}},\hat{\rho_{s}} } }{\min} | \rho - \mathbf{x} |1$,
where $\rho$ is the ground truth degree of polarisation (DoP) obtained from real polarisation images, and $\hat{\rho_{d}}$ and $\hat{\rho_{s}}$ are the analytically computed diffuse and specular DoP using the inverted physical model. The loss term aligns the predicted DoP of the student network with the ground truth DoP.
The overall loss is given by:
$\mathcal{L} = \mathcal{L}_{pseudo} + \mathcal{L}_{physics}$.

\section{Quantitative Results on Real Data}
\label{subsec:quantitative}
We conducted experiments to analyse the influence of input modality on pose estimation accuracy, specifically focusing on the impact of polarimetric image information. By comparing RGB-only and polar RGB inputs, we found that polarisation modality improves accuracy slightly, for the challenging object \textit{fork}. Incorporating shape information from polarimetric images enhances pose estimation significantly, as evidenced by improved normals prediction and overall performance. These results highlight the network's ability to establish a more accurate geometrical representation when guided by polarisation and shape cues (cf. Tab.~\ref{tab:ablation}).

Our experiments demonstrate the robustness of polarimetric imaging inputs in achieving accurate 6D pose prediction for photometrically challenging objects (cf. Tab.~\ref{tab:baseline:rgbd}). Compared to FFB6D~\cite{He_2021_CVPR}, which combines appearance and depth information, our method leverages polarimetric information to overcome challenges associated with photometric complexity, such as reflection or transparency. 

Table \ref{tab:full_model} presents quantitative results that demonstrate the effectiveness of our self-supervision pipeline. We evaluate the performance of the teacher and student networks under different conditions. The $\text{OURS}_\text{(Lower-Bound Teacher)}$ and $\text{OURS}_\text{(Lower-Bound Student)}$ denote the lower bound performance when trained on synthetic data and tested on real data. The $\text{OURS}_\text{(Upper-Bound Teacher)}$ and $\text{OURS}_\text{(Upper-Bound Student)}$ represent the upper bound performance when trained and evaluated on real data using ground-truth labels, where the $\text{OURS}_\text{(Upper-Bound Teacher)}$ is identical to PPP-Net~\cite{gao2021polarimetric}. Our model $\text{\textbf{OURS}}_\text{(RGB-Ps)}$ consistently outperforms the SOTA RGB-D method~\cite{wang2021occlusion} and achieves comparable results to the fully supervised upper bound baseline~\cite{gao2021polarimetric} for the \textit{cup} object, which is the least complex in terms of photometric considerations.

\begin{table}[!t]
\centering
\footnotesize
\caption{\textbf{Self-Supervised Quantitative Results. }Average recall of ADD(-S) metric is reported for different objects with increasing photometric complexity. $\dagger$ as upper bound is identical to PPP-Net~\cite{gao2021polarimetric}; Self6D++ from~\cite{wang2021occlusion}}
\resizebox{\linewidth}{!}{
\begin{tabular}{l|c|c|c|c|c}
\shline
\multicolumn{1}{c|}{Methods}  & \multicolumn{1}{c}{Cup} & \multicolumn{1}{c}{Fork} & \multicolumn{1}{c}{Knife} & \multicolumn{1}{c|}{Bottle} & \multicolumn{1}{c}{Mean}  \\

\hline
$\text{OURS}_\text{(Lower-Bound Student)}$ & 53.7 & 64.4 & 46.1 & 47.5 & 52.9 \\
$\text{OURS}_\text{(Lower-Bound Teacher)}$ & 72.3 & 75.0 & 67.3 & 76.2 & 72.7 \\
\hline
$\text{OURS}_\text{(Upper-Bound Student)}$ & 86.4 & 88.0 & 91.1 & 80.4 & 86.5 \\
$\text{OURS}_\text{(Upper-Bound Teacher)}$ $\dagger$ & 91.4 & 91.7 & 90.0 & 89.4 & 90.6 \\
\hline\hline
Self6D++ & 68.4 & 14.3 & 17.8 & 33.5 & 34.0 \\
$\text{\textbf{OURS}}_\text{(RGB-Ps)}$ & \textbf{93.8} & \textbf{72.4} & \textbf{78.4} & \textbf{78.2} & \textbf{80.7} \\

\shline
\end{tabular}
}

\label{tab:full_model}
\end{table}

\section{Conclusion}
Our proposed self-supervised framework leveraging polarimetric information has demonstrated its effectiveness in improving 6D object pose estimation accuracy. By integrating pseudo label learning and physics-induced self-supervision, our approach achieves superior performance compared to existing RGB-D methods and even approaches the results of fully supervised methods. The utilisation of polarimetric images as additional inputs provides valuable complementary information and helps bridging the domain gap between synthetic and real data. These findings highlight the potential of self-supervision and polarimetric imaging in advancing the field of object pose estimation.

\newpage

{\small
\bibliographystyle{ieee_fullname}
\bibliography{literature}
}

\end{document}